\documentclass[a4paper]{cas-sc}
\usepackage[authoryear]{natbib}
\usepackage{enumitem}
\usepackage{textcomp}
\usepackage{color}
\usepackage{amsmath}
\def\tsc#1{\csdef{#1}{\textsc{\lowercase{#1}}\xspace}}
\tsc{WGM}

\begin{document}
\let\WriteBookmarks\relax
\def\floatpagepagefraction{1}
\def\textpagefraction{.001}
\shorttitle{} 

\shortauthors{}
\title [mode = title]{Cross-Camera Cow Identification via Disentangled Representation Learning} 




%
\author[1]{Runcheng Wang}[type=editor, auid=000,bioid=1,style=chinese]



\ead{2024121171@sdau.edu.cn}


\credit{Conceptualization of this study, Methodology, Software}

\affiliation[1]{organization={College of Information Science and Engineering, Shandong Agricultural University},
    citysep={}, 
    postcode={271018},  
    city={Taian},
    state={},
    country={China}}

\author[2]{Yaru Chen}[style=chinese]
\affiliation[2]{organization={Centre for Vision Speech and Signal Processing (CVSSP), University of Surrey},
    postcode={Surrey GU2 7XH}, 
    country={United Kingdom}}

\author[1]{Guiguo Zhang}
\credit{Data curation, Writing - Original draft preparation}

\author[1]{Honghua Jiang}
\ead{cvr3@sayahna.org}
\cormark[1]

\credit{Data curation, Writing - Original draft preparation}

\author%
[3]
{Yongliang Qiao}
\cormark[2]
\ead{rishi@stmdocs.in}
\ead[URL]{www.stmdocs.in}

\affiliation[3]{organization={Australian Institute for Machine Learning, University of Adelaide},
    city={Malayinkil},
    citysep={}, 
    postcode={SA 5000}, 
    state={Adelaide},
    country={Australia}}

\cortext[cor1]{Corresponding author}
\cortext[cor2]{corresponding author}



\begin{abstract}
Precise identification of individual cows is a fundamental prerequisite for comprehensive digital management in smart livestock farming. While existing animal identification methods excel in controlled, single-camera settings, they face severe challenges regarding cross-camera generalization. When models trained on source cameras are deployed to new monitoring nodes characterized by divergent illumination, backgrounds, viewpoints, and heterogeneous imaging properties, recognition performance often degrades dramatically. This limits the large-scale application of non-contact technologies in dynamic, real-world farming environments. To address this challenge, this study proposes a cross-camera cow identification framework based on disentangled representation learning. This framework leverages the Subspace Identification Guarantee (SIG) theory in the context of bovine visual recognition. By simulating the underlying physical data generation process, we designed a principle-driven feature disentanglement module that decomposes the observed images into multiple mutually orthogonal latent subspaces. This model effectively isolates stable, identity-related biometric features that remain invariant across cameras, thereby substantially improving generalization to unseen cameras. We constructed a high-quality dataset spanning five distinct camera nodes, covering heterogeneous acquisition devices and complex variations in lighting and angles. Extensive experiments across seven cross-camera tasks demonstrate that the proposed method achieves an average accuracy of 86.0\%, significantly outperforming the Source-only Baseline (51.9\%) and the strongest cross-camera baseline method (79.8\%). This work establishes a feature disentanglement framework based on SIG theory for collaborative cross-camera cow identification, offering a new paradigm for precise animal monitoring in uncontrolled smart livestock environments.
\end{abstract}


\begin{highlights}
\item A disentangled representation learning framework based on subspace identification guarantee theory is proposed for cross-camera cow identification.
\item The model explicitly disentangles intrinsic biometric features from varying viewpoints and imaging styles.
\item A multi-view dataset (CCCI60) is constructed, where the method achieves superior generalization accuracy
\end{highlights}

\maketitle

\section{Introduction}
In precision livestock farming, the precise and efficient identification of individual cows is a fundamental prerequisite underpinning advanced applications such as behavior analysis, milk yield tracking, and precision feeding. It is crucial for enhancing animal welfare and achieving refined management \citep{awad2016classical}. Particularly within modern multi-camera collaborative monitoring systems, achieving seamless tracking of cows across the entire facility has emerged as a critical challenge for constructing fully automated, non-contact smart livestock.\newline\indent\setlength{\parindent}{2em} 
Non-invasive identification based on biometric features has become a prominent research focus. Among these, methods utilizing fine-grained features like iris patterns \citep{larregui2015biometric}, retinal vessel patterns \citep{saygili2024cattnis,cihan2025performance}, facial features \citep{weng2022cattle,xu2022cattlefacenet}, and muzzle prints \citep{li2022individual,shojaeipour2021automated} achieve high precision under controlled conditions. Nevertheless, their stringent requirements for acquisition distance and image quality hinder their widespread deployment in unconstrained breeding environments \citep{qiao2021intelligent}. In contrast, bovine body shape and black-and-white coat patterns, characterized by their ease of long-distance acquisition and strong robustness, have been widely adopted for building low-cost monitoring networks \citep{hossain2022systematic,kang2025research}. Currently, researchers utilize convolutional neural networks (CNNs) \citep{andrew2016automatic,andrew2017visual} to extract discriminative features from full-body posture, gait \citep{okura2019rgb}, and trunk patterns \citep{xiao2024novel,andrew2021visual}. Through hybrid attention architectures \citep{xu2024optimized} or improved metric learning \citep{zhao2022compact}, technological breakthroughs shifting from static recognition to dynamic tracking have been realized \citep{sim2025enhanced}.
However, existing identification methods commonly face a severe bottleneck regarding cross-camera generalization. This is attributed to significant physical discrepancies in illumination conditions, sensor parameters, and viewing angles, where perspective deformation and occlusion caused by varying heights and angles, combined with noise and color deviations from heterogeneous sensors and ambient light changes, collectively lead to significant data distribution shifts. Such inconsistency in physical imaging characteristics makes it difficult to directly transfer models trained on source cameras to unseen target cameras, while the prohibitive cost of data re-acquisition and fine-tuning for each new node severely obstructs the large-scale expansion of smart livestock technologies. \newline\indent\setlength{\parindent}{2em} 
To address the aforementioned bottleneck of cross-camera generalization, current mainstream solutions employ domain adaptive techniques, aligning feature distributions across different cameras by minimizing statistical distances in the feature space \citep{long2015learning} or introducing adversarial training mechanisms \citep{ganin2015unsupervised}. To handle complex shifts arising from multi-camera settings, researchers have further proposed multi-source adaptation strategies \citep{zhu2019aligning,li2021t}. More recent efforts have explored more refined approaches, such as utilizing cycle-inconsistency to suppress interference from irrelevant features \citep{lin2022adversarial}, or incorporating content-style disentanglement architectures \citep{liu2022learning}. In livestock identification scenarios, cumulative unsupervised adaptation has also been applied for knowledge accumulation across monitoring points \citep{dubourvieux2022cumulative}. However, these correlation-based alignment approaches remain limited in complex, uncontrolled farming environments: they essentially force the alignment of global distributions across different cameras while neglecting the underlying physical data generation process. Since cow identity information is deeply entangled with environmental factors, such coarse-grained alignment is susceptible to negative transfer phenomena. Specifically, while aligning background or lighting styles, it inadvertently corrupts the highly discriminative fine-grained biometric features of the cows, thereby limiting the model's generalization capability on unseen camera nodes.\newline\indent\setlength{\parindent}{2em} 
In recent years, disentangled representation learning has provided a novel theoretical perspective for resolving this generalization dilemma. Unlike traditional surface-correlation methods, this approach advocates explicitly modeling the underlying mechanisms of data generation, disentangling observed data into semantically relevant stable factors and environment-related varying factors \citep{kong2023partial,liu2025latent}. Advanced subspace identification theory demonstrates that by imposing appropriate structural constraints, models can identify and separate these latent feature subspaces from confounded visual signals, enabling accurate identification of target objects even in unseen stylistic scenarios \citep{li2023subspace}. This concept of feature disentanglement, grounded in underlying physical properties, provides a principled foundation for extracting interference-free intrinsic cow features within unstructured and complex farming environments.\newline\indent\setlength{\parindent}{2em} 
Motivated by the underlying physical imaging process and the Subspace Identification Guarantee (SIG) theory, this study move beyond traditional global feature alignment and propose a disentanglement-based model for cross-camera cow identification by explicitly modeling the underlying image formation process across different cameras.. This model is capable of deeply disentangling complex image representations, precisely extracting intrinsic, camera-invariant biometric features—typified by the unique topological structure of black-and-white coat patterns and body contours—while thoroughly disentangling view-dependent interference such as illumination patterns, viewing angles, and heterogeneous device discrepancies. This strategy circumvents the model's reliance on camera-specific attributes, thereby providing a principled mechanism for improved cross-camera generalization. To validate this approach, we constructed a high-quality multi-view dataset covering five distinct camera monitoring nodes and conducted extensive cross-camera experiments. Experimental results demonstrate that the proposed model achieves an average identification accuracy of 86.0\%, substantially outperforming the Source-Only baseline (51.9\%) and a strong multi-source adaptation method (iMSDA, 79.8\%). These results validate the effectiveness of disentangling identity-related representations from camera-induced variations in cross-camera cow identification.\newline\indent\setlength{\parindent}{2em} 
The main contributions of this study are as follows:

\begin{enumerate}[label=(\arabic*), leftmargin=4em]
\item We introduce the SIG theory into cross-camera cow identification, providing a principled perspective for addressing distribution shifts induced by multi-camera monitoring.
\item Based on the subspace identifiability for latent variables, we designed and implemented a deep learning model specialized for cross-camera cow identification. Addressing the uniqueness of bovine biometric features, we conducted an in-depth physical interpretation and utilization of the model's latent subspaces, explicitly disentangling visual features into imaging style factors and biological identity factors, thereby achieving the targeted elimination of interference information.
\item Through extensive experiments on a self-constructed cross-camera, multi-view cow dataset, we validated the effectiveness and superiority of the proposed disentanglement paradigm in agricultural vision tasks. Its performance significantly surpasses that of traditional single-camera methods and baseline adaptation approaches. This indicates that this approach has practical application potential in the multi-point collaborative monitoring of smart livestock.
\end{enumerate}  

\section{Materials and methods}
\subsection{ Cross-Camera Observation Nodes} 
Data were collected from April 6 to May 6, 2025, at a commercial dairy farm located in the Taishan District of Tai’an, China. Sixty lactating Holstein cows housed in Barn No. 6 were selected as subjects for long-term tracking. To construct a challenging cross-camera dataset that comprehensively reflects visual feature variations in a real-world farming environment, we strategically deployed a hybrid distributed acquisition system comprising six cameras along the mandatory route taken by cows from the barn to the milking parlor, with two cameras covering adjacent walking and milking corridors. This setup established five heterogeneous monitoring nodes characterized by significant discrepancies in illumination, viewpoint, and background distribution (as shown in Fig. 1).
\begin{figure}
	\centering
		\includegraphics[scale=.75]{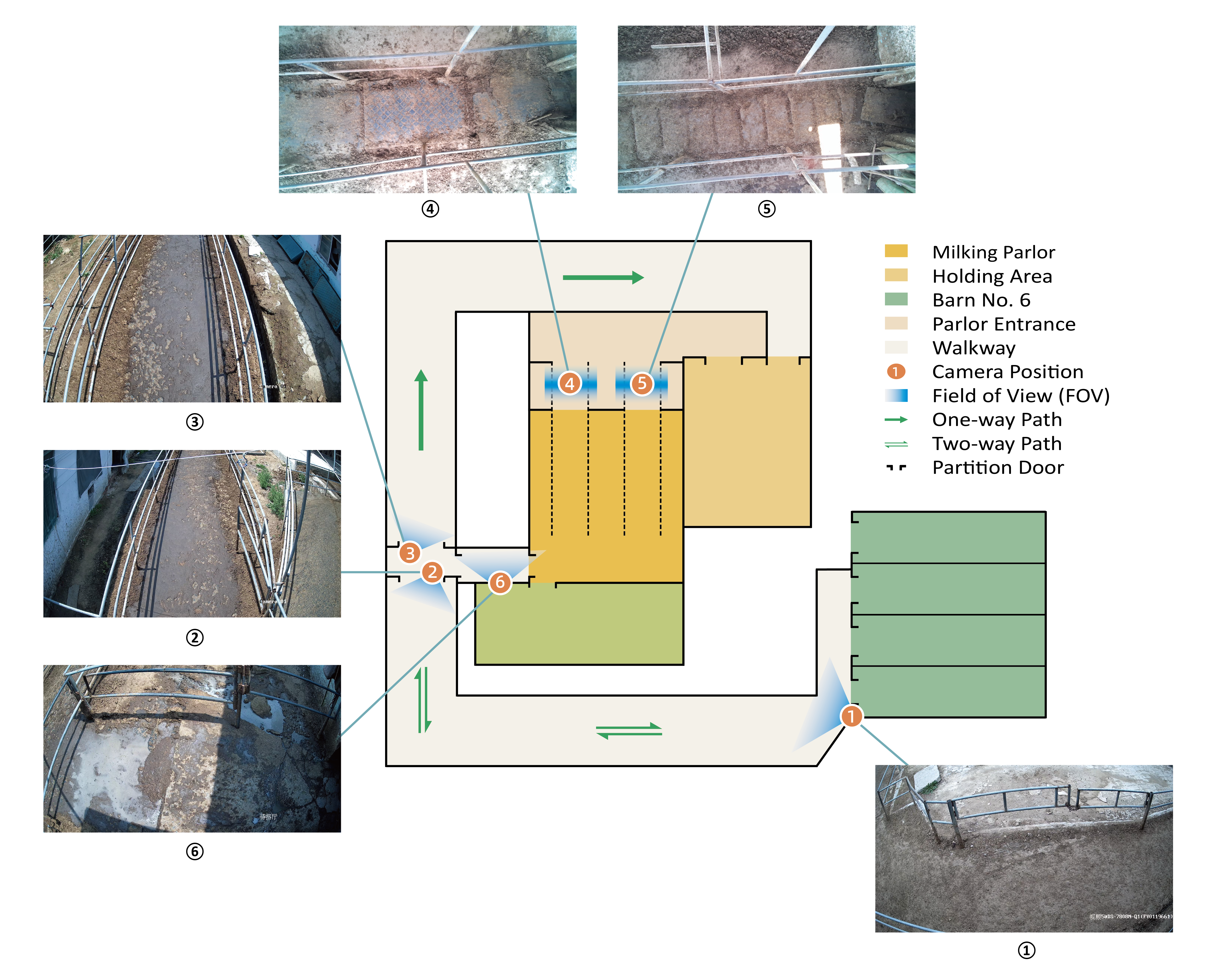}
	\caption{Layout of the cross-camera data acquisition system and scene examples. The central schematic illustrates the complete route of cows traveling from Barn No. 6 to the milking parlor and returning, where numbered circles \textcircled{1} through \textcircled{6} indicate the positions of the six cameras. The surrounding panels \textcircled{1} through \textcircled{6} display real-world scene images captured by the cameras at the corresponding positions. The specific layout is as follows: \textcircled{1} Barn Exit; \textcircled{2} and \textcircled{3} Walking Aisle; \textcircled{4} and \textcircled{5} Milking Parlor Entrance; and \textcircled{6} Milking Parlor Exit.}
	\label{FIG:1}
\end{figure}\newline\indent\setlength{\parindent}{2em} 
This multi-view deployment strategy is designed to expose substantial appearance variations of cows across different physical spaces. By introducing heterogeneous imaging devices and variable physical environments (e.g., illumination deviations, perspective distortions), we compel the model to strip away camera-related interference factors during the learning process and focus on the intrinsic biometric features of individual cows.\newline\indent\setlength{\parindent}{2em} 
Specifically, at the Barn Exit , one network camera was deployed to capture standard lateral views of cows leaving the barn. This node features a relatively open field of view with sufficient and uniform lighting, enabling the acquisition of high-quality, information-rich images of individual cows.\newline\indent\setlength{\parindent}{2em} 
In the Walking Aisle connecting the barn and the milking parlor, we deployed two network cameras from opposite directions to capture frontal and rear views of the moving cows, respectively. The frontal and rear viewing angles cause severe perspective foreshortening of the trunk patterns. Coupled with the muscular movement during walking, the body surface exhibits complex non-rigid deformations. \newline\indent\setlength{\parindent}{2em} 
At the Milking Parlor Entrances of two parallel milking lanes, we installed one depth camera each, both utilizing a top-down view for recording. The mounting parameters (height, angle) of the two cameras were kept consistent. In this section, the field of view is limited, and the cows are crowded; the top-down images mainly contain pattern information from the cows' backs. \newline\indent\setlength{\parindent}{2em} 
Finally, at the Milking Parlor Exit where cows return after milking, we deployed another network camera to capture lateral views. This scene is equipped with supplementary lighting that activates when ambient light is insufficient, thereby introducing significant artificial lighting interference. \newline\indent\setlength{\parindent}{2em} 
In the aforementioned monitoring network, to further introduce diversity in imaging styles, we employed two types of heterogeneous acquisition devices. The monitoring system comprises four  network cameras (Model DS-2CD3T67WDV3-L, Hikvision, Hangzhou, China) and two depth cameras (Model Azure Kinect DK, Microsoft, Redmond, USA). Although all nodes uniformly utilize RGB image streams as input to maintain modality consistency, significant discrepancies exist in the underlying hardware specifications between the Hikvision and Azure Kinect devices. As shown in Table 1, the two devices exhibit clear differences in key imaging parameters, such as optical format, dynamic range, and pixel size. These disparities constitute a major source of imaging style shift in cross-camera identification.

\begin{table}[width=.5\linewidth,cols=3,pos=h]
\caption{Detailed comparison of hardware specifications for heterogeneous imaging devices}\label{tbl1}
\begin{tabular*}{\tblwidth}{@{} LLL @{} }
\toprule
Specification & Azure Kinect & Hikvision\\
\midrule
Sensor Type & OV12A10 CMOS & Progressive Scan CMOS \\
Optical Format & 1/2.8 " & 1/2.4 " \\
Dynamic Range & 75 dB & 120 dB\\
Pixel Size & 1.25 $\mu$m & 1.4 $\mu$m \\
\bottomrule
\end{tabular*}
\end{table}
\subsection{Datasets: CowId60} 
We construct a dataset named CCCI60, comprising 7,378 finely annotated images of 60 lactating Holstein cows captured by a multi-view camera monitoring system. As shown in Table 2, the samples are distributed across 5 key physical monitoring scenarios, including the barn exit, walking aisle, and milking parlor entrance, forming a challenging benchmark for cross-camera evaluation. A defining characteristic of this dataset is the coexistence of drastic intra-class variation and clear inter-class differences. Under the combined influence of varying illumination modes, viewing angles, and non-rigid pose changes during cow movement across different monitoring nodes, the appearance images of the same cow exhibit significant style shifts across cameras. This leads to a deep entanglement of the cow's intrinsic biometric features with the noise of the variable imaging environment. As illustrated in Fig. 2, although the switching of monitoring nodes induces drastic fluctuations in image texture representation, the unique topological structure of black-and-white coat patterns among different individuals still demonstrates strong physical stability. This visual complexity across scenes and viewpoints provides solid data support for evaluating whether the model can disentangle camera-invariant intrinsic identity representations from confounded visual signals, thereby achieving precise identification in uncontrolled breeding environments.

\begin{table}[width=.5\linewidth,cols=3,pos=h]
\caption{Statistical distribution of image samples across different physical scenarios in the constructed cross-camera dataset.}\label{tbl2}
\begin{tabular*}{\tblwidth}{@{} LLL@{} }
\toprule
Camera & Location Description & Number of Images\\
\midrule
1 & Barn Exit & 857 \\
2 & Walking Aisle (Backward) & 1052 \\
3 & Walking Aisle (Forward) & 1672 \\
4 & Milking Parlor Entrance & 2157 \\
5 & Milking Parlor Exit & 1640 \\
Total & All 5 Scenarios & 7378\\
\bottomrule
\end{tabular*}
\end{table}

\begin{figure}
	\centering
		\includegraphics[scale=.36]{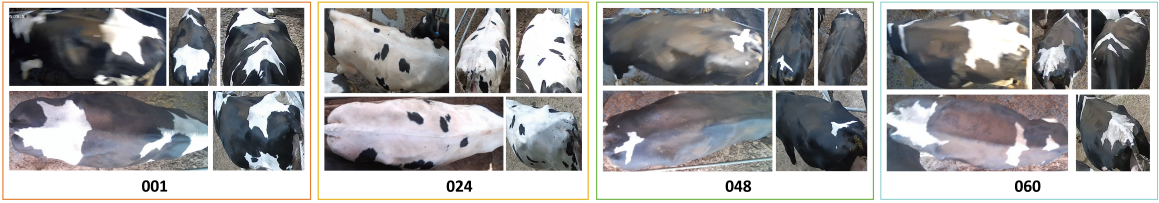}
	\caption{Visualization of samples from the cross-camera cow identification dataset. The figure displays representative images of four different cows (ID: 001, 024, 048, 060).}
	\label{FIG:2}
\end{figure}

\subsection{Problem formulation and physical generative logic} 
In smart livestock monitoring networks, cow images collected from different cameras often exhibit significant discrepancies at both the pixel and feature levels. These discrepancies arise from uneven illumination, varying camera installation angles, and background fluctuations, and they substantially degrade identification performance. To address this issue, this study formulates the task as a multi-source domain adaptation problem.\newline\indent\setlength{\parindent}{2em} 
Assume there are $K$  source cameras with annotated images deployed in the farm, denoted as the set $S = \left\{ {{S_1},{S_2}, \ldots ,{S_K}} \right\}$ , and one unlabeled target camera ${S_T}$ . For any camera $u \in S \cup {S_T}$ , the collected cow image $x\;$ and its identity label $y\;$ follow a joint distribution $p\left( {x,y\mid u} \right)$ . The objective is to utilize the labeled data from the source camera set $S$ , denoted as  ${D_S} = \{ \left( {{x_i},{y_i},{u_i}} \right)\} _{i = 1}^{{N_S}}$, to achieve precise prediction of the distribution $p\left( {y\mid x,{u_{{S_T}}}} \right)\;$ on the unseen target camera ${S_T}$  via disentangled  intrinsic identity features. However, due to the heterogeneity of the monitoring environment, significant shifts exist in the marginal distributions between different cameras, i.e., $p\left( {x\mid {u_i}} \right) \ne p\left( {x\mid {u_j}} \right)$ . This causes the generalization capability of identification models trained on source cameras to drop sharply when directly transferred to the target camera.\newline\indent\setlength{\parindent}{2em} 
The goal of this study is to map the image $x\;$ to a high-dimensional latent variable space $z\;$ through a feature disentanglement mechanism. By identifying intrinsic features capable of representing individual identity from this space, the identity-discriminative patterns learned on source cameras can be directly transferred to the target camera, thereby adapting to practical deployment scenarios involving multiple cameras.

\begin{figure}
	\centering
		\includegraphics[scale=.36]{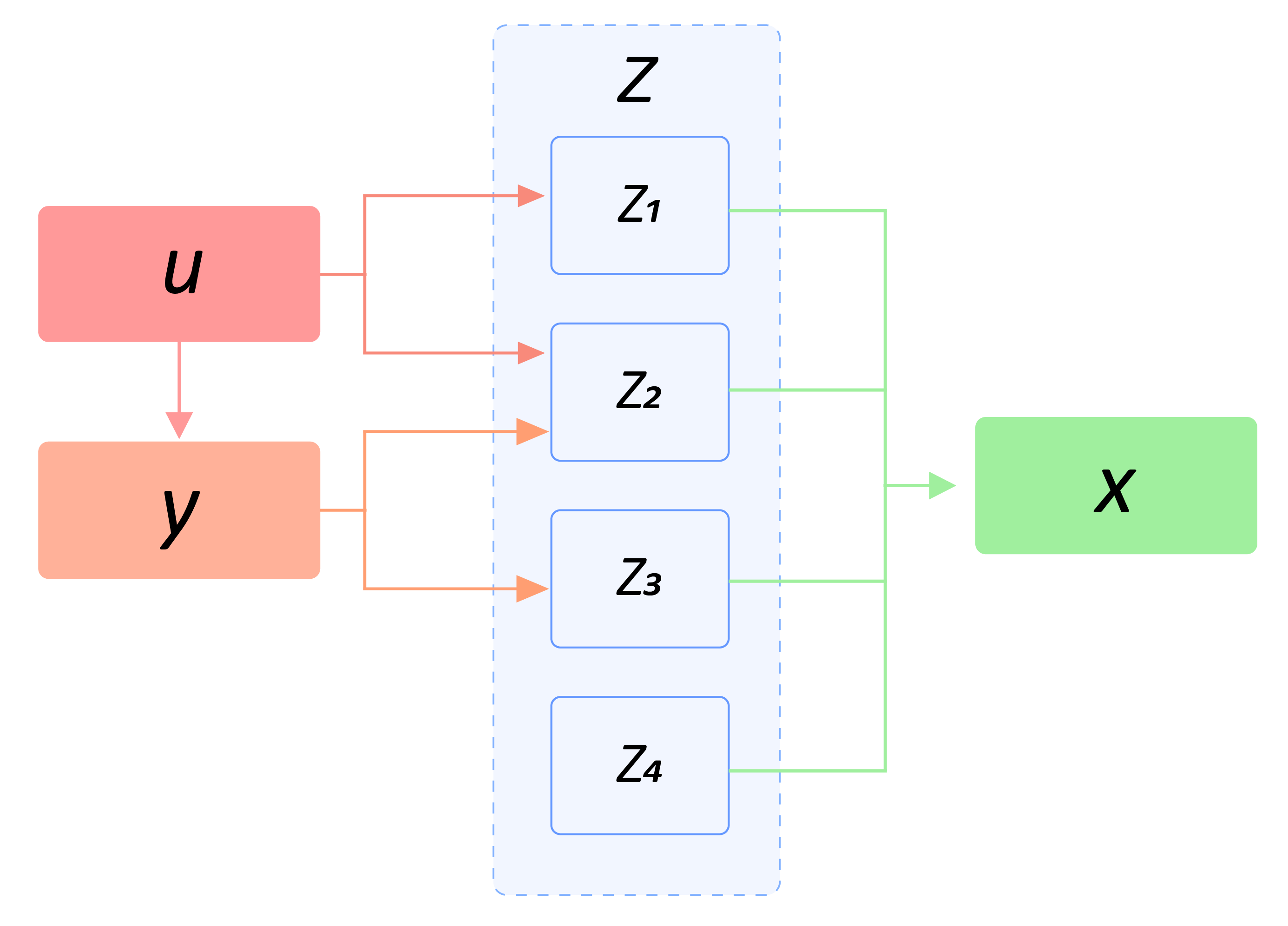}
	\caption{Physical generative graph. Observed variables: Camera index $u$ , individual identity label $y$ , and cow image $x$ . Latent variables $z = \left\{ {{z_1},{z_2},{z_3},{z_4}} \right\}$ }
	\label{FIG:3}
\end{figure}

\indent\setlength{\parindent}{2em}
We hypothesize that the generation process of cow images $x\;$ observed in the dynamic breeding environment follows the physical generative graph shown in Fig. 3. Accordingly, we map the image $x\;$ to a latent space $z$ and explicitly partition it into four orthogonal subspaces, i.e., $z = \left\{ {{z_1},{z_2},{z_3},{z_4}} \right\}$ , defined as follows:

\begin{itemize} 
\item Camera-specific imaging style variables $z_{1}\in \mathbb{R}^{d_{1}}$.
\item View-dependent biometric feature variables $z_{2}\in\mathbb{R}^{d_{2}}$.
\item Intrinsic biometric topology variables $z_{3}\in\mathbb{R}^{d_{3}} $.
\item Universal species appearance variables $z_{4}\in\mathbb{R}^{d_{4}} $.
\end{itemize}

To provide a concrete physical interpretation of these latent variables within the context of precision livestock farming, we analyze their distinct semantic roles. First, the directed edge from camera index $u$ to identity label $y$ in the graph models the label shift phenomenon, reflecting how the occurrence frequency of cows varies across different functional zones. Regarding the latent subspaces, ${z_1}$ aims to encode information solely determined by heterogeneous acquisition devices (e.g., sensor parameters, lens distortion) and environmental lighting, which serves as interference irrelevant to individual identity. ${z_2}$  captures the interaction effect between individual biometric features and camera viewing angles, characterizing specific manifestations such as the geometric stretching or compression of coat patterns due to varying viewpoints. In contrast, ${z_3}$ contains the essential biological markers possessing cross-camera invariance, specifically the unique topological structure of black-and-white coat patterns and the connectivity between patches, which serves as the fundamental basis for reliable identification. Finally, ${z_4}$ represents visual attributes common to all Holstein cows that do not change with camera viewpoints, such as the basic coat color distribution; this information is redundant for identity discrimination and should be disentangled from individual-specific features.

\subsection{Methodology overview} 
The overall identification workflow proposed in this study is illustrated in Fig. 4, encompassing three primary stages: preprocessing, feature disentanglement, and conditional decision-making. In the first stage, standardized cow trunk image sequences are detected and extracted from multi-view video streams utilizing the YOLOv11n object detection algorithm. In the second stage, following the physical generative logic defined in Section 2.3, visual representations are explicitly mapped into multiple latent subspaces, achieving the separation of environmental interference from intrinsic biometric features. In the third stage, the decision process is conditioned on the camera index u together with the disentangled view-dependent features ${z_2}$ and intrinsic biometric features ${z_3}$, enabling adaptive utilization of auxiliary discriminative cues under specific observation conditions. This pipeline, through the precise disentanglement and integration of entangled signals, effectively safeguards the system's identification robustness across varying viewpoints and cameras.

\begin{figure}
	\centering
		\includegraphics[scale=.57]{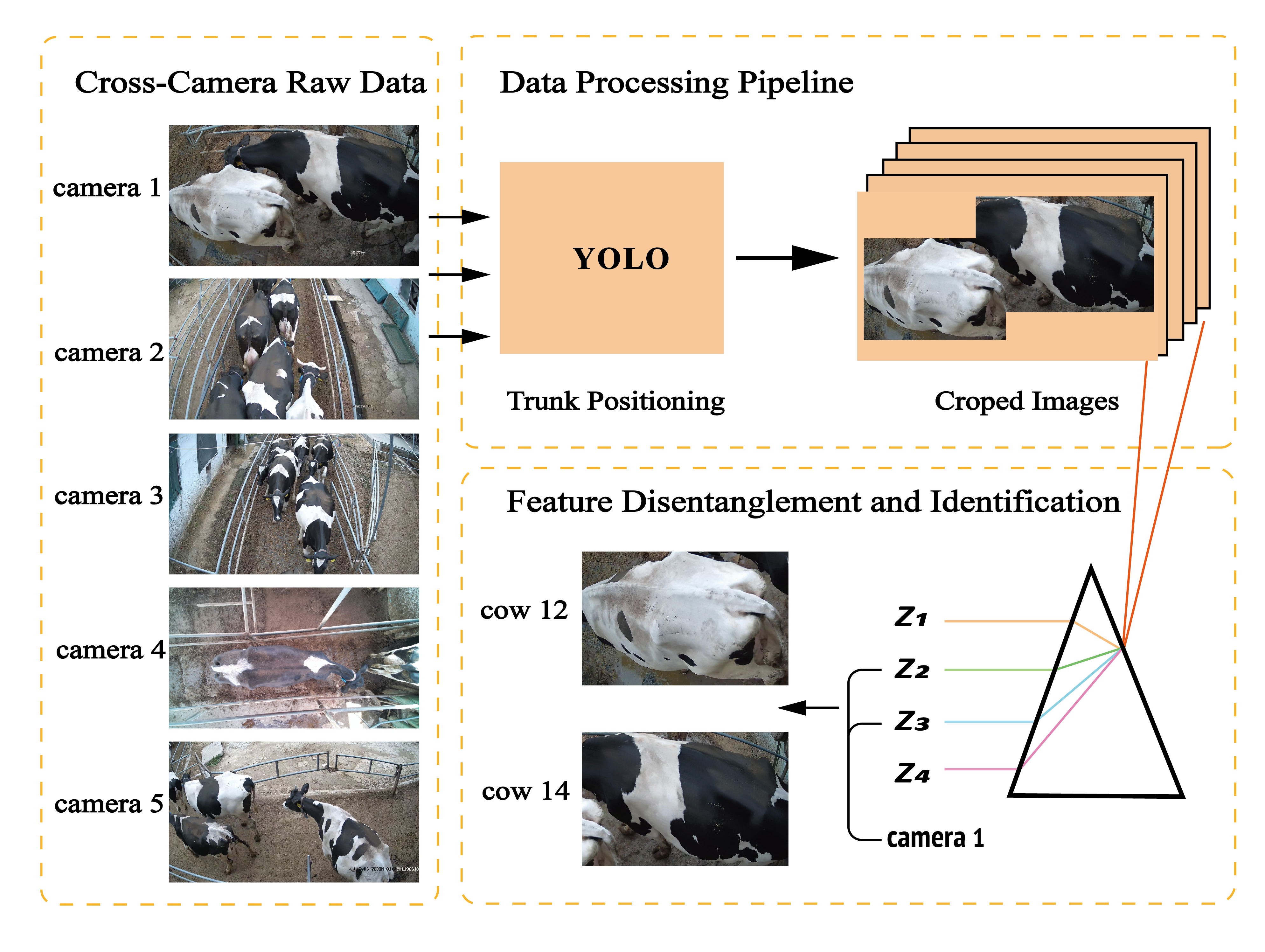}
	\caption{Schematic overview of the proposed framework}
	\label{FIG:4}
\end{figure}

\subsection{Automated cattle trunk extraction}
In high-density farming environments, surveillance video footage typically presents a complex state of multi-target coexistence, which is not directly compatible with the specific requirement of single-target input for individual identification algorithms. To address this, the first stage of our workflow employs the YOLOv11n object detection algorithm to construct an automated processing pipeline, to precisely localize and extract individual cows from raw video frames. This module focuses the detection bounding box specifically on the cow's trunk region for two primary reasons: First, as the body part with the largest surface area, the trunk contains the richest topological features of black-and-white patterns, exhibiting the strongest discriminability and stability under cross-camera viewpoints. Second, in crowded herd environments, localizing the trunk—where features are most prominent—enables more effective separation of independent single-cow samples from visually cluttered backgrounds. Through this automated extraction process, the system successfully transforms raw video streams into discrete sequences of single-body images with clear biometric features and uniform specifications, providing a standardized data foundation for the subsequent deep feature disentanglement stage.

\subsection{Implementation of visual feature disentanglement and conditional identification architecture}

The identification architecture proposed in this study aims to disentangle feature with clear physical connotations from confounded visual observations, based on the SIG theory \citep{li2023subspace}. As shown in Fig. 5, the architecture consists of three interconnected components: feature mapping, disentanglement constraints, and conditional decision-making.

\begin{figure}
	\centering
		\includegraphics[scale=.3]{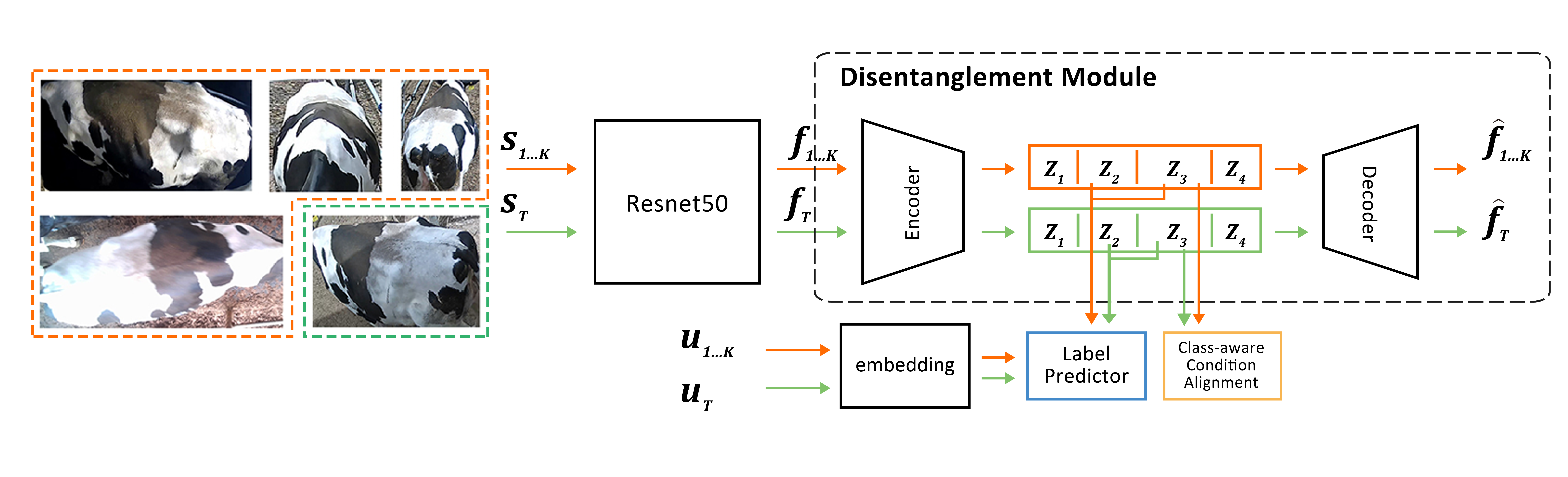}
	\caption{Architecture of the proposed model}
	\label{FIG:5}
\end{figure}

\subsubsection{Deep feature mapping and probabilistic modeling}
The model employs a pre-trained ResNet-50 network as the feature extraction backbone. This module accepts preprocessed individual cow images as input, mapping them into a high-dimensional space to obtain feature vectors $f$ . Subsequently, the system utilizes a variational encoder based on multilayer perceptron (MLP) to further map the features $f$ to the mean and variance of the latent variable space $z$ , employing the reparameterization trick. This step transforms raw pixel representations into latent representations with probabilistic attributes, providing the mathematical foundation for the fine-grained disentanglement.

\subsubsection{Logic of disentanglement constraints on latent subspaces}
Following the physical generative logic defined in Section 2.3, the latent variable space $z$ is explicitly partitioned into four mutually orthogonal subspaces. To guarantee the independence of each subspace and the integrity of discriminative information, two key categories of physical logic constraints are introduced into the architecture design. On one hand, the model reconstructs deep features from the latent variables $z$ via a decoder module. The optimization process of the Evidence Lower Bound (ELBO) ensures that all subspaces jointly cover the complete physical information of the cow images, effectively preventing the loss of critical biological features during disentanglement. On the other hand, an auxiliary prediction branch is constructed to predict the camera index $u$ from non-discriminative subspace information such as imaging style. This targeted constraint logic compels all style interference factors arising from camera node switching to be precisely compressed into specific subspaces, thereby achieving a deep separation of environmental noise from intrinsic identity features.

\subsubsection{Construction of the conditional joint decision-making mechanism}
The final identity prediction logic of this study is founded on the joint modeling of view-dependent features ${z_2}$ , intrinsic biometric features ${z_3}$ , and the camera index $u$ . The classification decision head integrates these three types of physical factors with distinct attributes to construct a conditional identification mechanism capable of coping with varying viewpoint challenges. Specifically, the model performs dimension concatenation of the disentangled biological feature subspaces and the camera index information. In this architecture, the camera index $u$ serves as a physical context reference for the decision-making process, guiding the model to adaptively extract auxiliary local details from view-dependent features ${z_2}$  under different observation angles and fuse them with cross-camera invariant intrinsic biometric topological features ${z_3}$ . This joint decision scheme greatly enhances the system's discrimination precision for individual cows under extreme viewpoints and complex backgrounds, while simultaneously ensuring the model's cross-camera transfer stability.

\subsection{Optimization objectives}
Based on the physical generative mechanism and SIG theoretical framework proposed in Section 2.3, the total loss function is formulated as a weighted combination of three core components: the identity classification module, the feature disentanglement module, and the feature alignment module. Formally, the total optimization objective is defined as: 
\begin{equation}
\mathcal{L}_{\text{total}} = \mathcal{L}_{\text{identity}} + \lambda_{\text{dis}} \mathcal{L}_{\text{disentangle}} + \lambda_{\text{align}} \mathcal{L}_{\text{align}}
\end{equation}
where ${\lambda _{dis}}\;$ and ${\lambda _{align}}\;$ are hyperparameters used to balance the contribution of the feature disentanglement and feature alignment constraints, respectively. In the following subsections, we will elaborate on the theoretical derivation and specific algorithmic implementation of each component.

\subsubsection{Identity classification module ($\mathcal{L}_{\text{identity}}$)}
(1) Weighted conditional classification loss for label shift \newline\indent\setlength{\parindent}{2em}
Although this study utilizes labeled source camera data to train the classifier, the model faces a severe challenge of label shift in practical smart livestock environments. Since different camera nodes cover distinct functional zones (e.g., barns, walking aisles, and milking parlors), the occurrence frequency distribution p(y) of the same cow varies significantly across monitoring points. Ignoring such differences in class prior distributions inevitably leads to negative transfer during feature alignment. In such cases, if standard cross-entropy loss is directly adopted, the model will tend to overfit classes with higher occurrence frequencies on the current camera, thereby severely impairing the identification performance for low-frequency classes on the target camera. To fundamentally correct this statistical bias induced by physical spatial distribution differences and to ensure the model adheres to the physical generative logic defined in Section 2.3, this study proposes a dynamic re-weighting conditional classification mechanism.\newline\indent\setlength{\parindent}{2em}
Specifically, the disentangled intrinsic biometric topological features ${z_3}$ , view-dependent features ${z_2}$ , and the camera index $u$ are concatenated and fed into a label predictor to model the conditional class probability. To compensate for label distribution discrepancies across different camera nodes, we introduce a class importance weight vector ${\alpha _u} \in {R^K}$  into the loss formulation. For a sample $({x_i},{y_i})$ in the  $u$-th camera, the weighted conditional classification loss is defined as:
\begin{equation}
\mathcal{L}_{cls} = -\frac{1}{N_S} \left( \sum_{i=1}^{N} \sum_{k=1}^{K} \alpha_{i,k} \cdot \mathbf{1}(y_i = k) \log p(y_i = k \mid z_{2,i}, z_{3,i}, u_i) \right)
\end{equation}
where $K$ represents the total number of cow individuals, and  $1\left(  \cdot  \right)$ denotes the indicator function. ${\alpha _{u,k}}$ denotes the re-weighting coefficient for the  $k$-th class of cows under camera $u$ . This coefficient is dynamically updated via the Black Box Shift Estimation (BBSE) strategy:
\begin{equation}
\alpha_{i,k} = \frac{\hat{p}_T(y = k)}{p_{S_u}(y = k)}
\end{equation}
where ${p_{{S_u}}}\left( y \right)$ is the known label distribution of the source camera, and ${\hat p_T}\left( y \right)$ is the estimated pseudo-label distribution predicted by the model on the target camera. Through this re-weighting strategy, the model is forced to pay attention to individuals that may be important in the target camera but are scarce in the current source camera, thereby eliminating the label shift introduced by locational differences.\newline\indent\setlength{\parindent}{2em}
(2) Minimum Class Confusion Loss\newline\indent\setlength{\parindent}{2em}
Although the conditional classification loss ${{\cal L}_{cls}}$ ensures the model's discriminative ability on source cameras, when facing unlabeled target cameras, the model is prone to producing ambiguous prediction results due to drastic changes in illumination, viewpoints, and sensor parameters. To enhance the model's adaptability under unseen cameras and reinforce inter-class boundaries, we introduce the Minimum Class Confusion (MCC) loss. Unlike traditional Entropy Minimization, which may lead to model collapse into a single class, MCC aims to minimize the pairwise correlations between different cow identities, thereby promoting distinct separability of features in the target feature space.\newline\indent\setlength{\parindent}{2em}
Specifically, given a batch of samples ${B_T}$  from the target camera, the model produces a logit vector${l_i} \in {l^K}\;$ for each sample. First, to smooth the prediction distribution and obtain more robust probability estimates, we introduce a temperature parameter $T$ to scale the output, obtaining the probability ${\hat y_{i,j}}\;$ that the target sample belongs to the $j$-th identity class:
\begin{equation}\hat{y}_{i,j} = \frac{\exp(l_{i,j} / T)}{\sum_{k=1}^K \exp(l_{i,k} / T)}\end{equation}
To suppress the negative impact of noisy samples on optimization, this framework adopts an entropy-based weighting mechanism. By defining the predicted entropy $H\left( {{{{\rm{\hat y}}}_i}} \right)$  of sample $i$  and introducing a small constant $\varepsilon$  (set to $1e - 5$ ) to ensure numerical stability:
\begin{equation}
    H(\hat{\mathbf{y}}) = - \sum_{j=1}^K \hat{y}_{i,j} \log\left(\hat{y}_{i,j} + \epsilon\right)
\end{equation}
Based on this entropy value, we calculate a weight ${w_i}$ for each sample, assigning higher weights to high-confidence samples. To maintain numerical consistency within the batch, we further apply batch-wise normalization to the weights:
\begin{equation}
    W_i = \frac{w_i}{\sum_{m=1}^{|B_T|} w_m} \cdot |B_T|,\quad \text{where } w_i = 1 + \exp(-H(\hat{\mathbf{y}}_i))
\end{equation}
Subsequently, the model constructs the initial weighted class correlation matrix ${C^{raw}} \in {R^{K \times K}}$, where the element $C_{jk}^{raw}\;$  represents the co-occurrence strength between the $j$-th class and the $k$-th class in the current batch:
\begin{equation}
    C_{j k}^{r a v}=\sum_{i=1}^{\left|B_{T}\right|}\left(\left(\hat{y}_{i, j} \cdot w_{i}^{-}\right) \cdot \hat{y}_{i, k}\right)
\end{equation}
To eliminate the absolute numerical bias caused by the imbalanced number of samples across different classes within the batch, the matrix  ${C^{raw}}$ undergoes row-wise normalization to yield the final confusion matrix $C$ :
\begin{equation}
    C_{j k}=\frac{C_{j k}^{r a w}}{\sum_{m=1}^{K} C_{j m}^{r a w}}
\end{equation}
Finally, the  ${{\cal L}_{mcc}}$ loss penalizes inter-class confusion by suppressing the sum of the off-diagonal elements, thereby encouraging the feature distribution of the target camera to be more compact and well-separated at the identity level:
\begin{equation}
    \mathcal{L}_{mcc} = \frac{1}{K} \left( \sum_{j=1}^K \sum_{k=1}^K |C_{j,k}| - \text{Tr}(C) \right)
\end{equation}
where ${\rm{Tr}}\left( C \right)$ denotes the trace of the matrix. By optimizing ${{\cal L}_{mcc}}$, the model can leverage unlabeled data to explicitly widen the distance between different cow individuals in the feature space, thereby achieving robust identity clustering under the target camera even in the absence of supervision labels.
\subsubsection{Feature Disentanglement Module $\mathcal{L}_{\text{disentangle}}$}
To extract robust identity features from complex observation images, we designed a Variational Autoencoder (VAE) inspired on the SIG theory. The objective of this module is not merely simple image reconstruction, but rather to explicitly disentangle the latent feature space $z$ into four subspaces with distinct physical semantics through mandatory structural constraints: $z = \left\{ {{z_1},{z_2},{z_3},{z_4}} \right\}$. These correspond to the camera-specific imaging style, view-dependent biometric features, intrinsic biometric  features, and universal species appearance, respectively.\newline\indent\setlength{\parindent}{2em}
The optimization process of the feature disentanglement module is jointly driven by variational reconstruction constraints and camera-index-based structural constraints. First, to ensure that the latent variables can fully preserve the biological and environmental information from the original images, the model maps the input images to a latent distribution via the encoder ${E_\phi }$ and performs feature reconstruction using the decoder ${D_\theta }$. By optimizing the Evidence Lower Bound (ELBO), the model minimizes the reconstruction error while constraining the posterior distribution to approximate a standard normal prior. The specific variational generative loss ${{\cal L}_{vae}}$ is defined as:
\begin{equation}
    \mathcal{L}_{vae} = \| x - D_\theta(z) \|_2^2 + \beta \cdot \left| D_{KL}\left(q_\phi(z \mid x) \| p(z)\right) - C \right|
\end{equation}
where the first term, the reconstruction error, ensures the integrity of discriminative information, and the second term, the KL divergence, is used to constrain the statistical distribution of the latent variable space. A progressive capacity threshold $C$ is incorporated to prevent posterior collapse and enhance representation quality. However, relying solely on unsupervised reconstruction makes it difficult to guarantee that the subspaces are partitioned according to their physical semantics. According to subspace identification theory, the camera index $u$ is a key conditioning variable associated with systematic variations in imaging style ${z_1}$ and view-dependent posture ${z_2}$. Therefore, this architecture introduces an auxiliary predictor ${P_\psi }$, which utilizes a combination of latent variables $\left\{ {{z_1},{z_2},{z_3}} \right\}$ to explicitly predict the camera node to which the image belongs. Its corresponding structural constraint loss ${{\cal L}_{dom}}\;$  is expressed as:
\begin{equation}
    \mathcal{L}_{dom} = - \sum_{m=1}^M \mathbb{1}(u = m) \log P_\psi(u = m \mid z_1, z_2, z_3)
\end{equation}
By minimizing ${{\cal L}_{dom}}$, the model is compelled to utilize camera index information to anchor the physical distribution structure of the latent variables. This directional constraint mechanism effectively compresses camera-varying interferences into specific subspaces, preventing them from leaking into subspace ${z_3}$. Consequently, feature-level disentanglement of bovine biometric traits is achieved.
\subsubsection{Feature Alignment Module ($\mathcal{L}_{\text{align}}$)}
To mitigate the impact of cross-camera label distribution shift, this study introduces a class-aware centroid alignment mechanism. This mechanism is designed to resolve the distribution mismatch caused by the varying occurrence frequencies of cows at different camera nodes. By explicitly calibrating the feature centroids of the same cow individual across the source and target cameras, the model is forced to learn a cross-camera invariant conditional feature distribution for specific individuals within the intrinsic identity subspace.\newline\indent\setlength{\parindent}{2em}
In implementation, the system employs a dynamic centroid estimation strategy to maintain the stability of global features. For the source camera $u$, the feature centroid for each cow class is calculated using ground-truth labels; for the unlabeled target camera $T$, the centroids are estimated using pseudo-labels $y$ generated by the classifier. To reduce random fluctuations during the training process, an Exponential Moving Average (EMA) strategy is adopted for smooth updates of the global centroids. Let $\mu _{u,k}^{\left( t \right)}$ and $\mu _{T,k}^{\left( t \right)}$ denote the feature centroids of the $k$-th cow class in the source camera $u$ and target camera $T$ at the $t$-th iteration, respectively. The update rules are as follows:
\begin{equation}
    \mu_{i,k}^{(t)} = \gamma \cdot \mu_{i,k}^{(t-1)} + (1 - \gamma) \cdot \bar{z}_{i,k}^{(t)}
\end{equation}
\begin{equation}
    \mu_{T,k}^{(t)} = \gamma \cdot \mu_{T,k}^{(t-1)} + (1 - \gamma) \cdot \bar{z}_{T,k}^{(t)}
\end{equation}
where $\bar{z}_{i,k}^{(t)}$ represents the mean feature of the $k$-th class samples observed within the current batch, and $\gamma$ is the momentum coefficient. Upon obtaining stable centroid estimates, the system defines the semantic alignment loss ${{\cal L}_{align}}$ based on the weighted Euclidean distance between the centroids of the two cameras. To address the label distribution shift issue discussed in Section 2.7.1, class importance weights ${\alpha _{u,k}}$ are re-incorporated into the alignment process. This strategy ensures that the optimization process prioritizes individuals with higher occurrence frequencies in the target camera, thereby effectively preventing the model from being dominated by scarce or noisy classes.
\begin{equation}
    \mathcal{L}_{align} = \sum_{k=1}^K \alpha_{u,k} \cdot \|\mu_{u,k} - \mu_{T,k}\|_2^2
\end{equation}
By minimizing ${{\cal L}_{align}}$, the model explicitly aligns the feature centroids of the same identity across different observation scenarios, ensuring reliable cross-camera identification at the feature distribution level.
\section{Experimental Setup}
\subsection{Training Dataset Construction}
\subsubsection{Object detection model training and trunk extraction}
To precisely isolate the identification targets from the raw monitoring videos containing multiple individuals, this study first trained a YOLOv11n object detection model to achieve automated trunk localization. We uniformly sampled 3,088 representative images from the collected 22,397 video frames to serve as dedicated training data for the detection model, covering five monitoring scenarios and various lighting conditions. During the data annotation phase, the bounding boxes were primarily focused on the trunk region of the cows, aiming to achieve individual separation in crowded herd environments. This annotated subset was randomly partitioned into training and validation sets at a 9:1 ratio for parameter optimization. The trained model was then deployed across all raw images to automatically output the coordinates of each cow, providing a standardized data foundation for the subsequent feature disentanglement stage.
\subsubsection{Object detection model training and trunk extraction}
This study designed a semi-automated data cleaning strategy combining temporal association with manual verification. First, taking advantage of the high frame rate of the video, preliminary individual association was achieved by calculating the spatial overlap of detection boxes in adjacent frames. Specifically, the Dice Coefficient was used as the evaluation metric, with its calculation formula defined as:
\begin{equation}
    Dice(A,B) = \frac{2|A \cap B|}{|A| + |B|}
\end{equation}
The system set a preset threshold of 0.8; only when the Dice Coefficient of detection boxes in adjacent frames exceeds this value are they determined to belong to the same moving target and automatically aggregated into a continuous trajectory. Subsequently, manual scrutiny and screening were performed on the generated image sequences based on three rigorous core criteria. First is the integrity criterion, which involves removing images where the visible trunk area is less than 80\% due to fence occlusion or moving to the edge of the field of view, ensuring that samples contain complete biometric topological structures. Second is the clarity criterion, which removes blurred images caused by rapid cow movement or camera focus failure. Finally, the diversity criterion was applied to manually eliminate redundant samples that are temporally adjacent and possess extremely similar visual features, thereby reducing the autocorrelation of training data and preventing overfitting.\newline\indent\setlength{\parindent}{2em}
After completing the aforementioned screening, this study performed a logical verification of identity labels on the remaining samples, correcting potential identity switching errors that occurred during the multi-object tracking process to ensure the accuracy of the ground-truth labels. Following this processing pipeline, the final CCCI60 dataset was constructed, comprising 7,378 high-quality images of 60 individual cows. Detailed sample statistics for each monitoring scenario are presented in Table 2.
\subsection{Statistical Verification of Cross-Camera Distribution Shift}
To verify the objective existence and significance of the distribution shifts induced by different cameras from a statistical perspective, this study introduced the Maximum Mean Discrepancy (MMD) as a non-parametric hypothesis testing tool \citep{borgwardt2006integrating,gretton2012kernel}. MMD aims to quantify the discrepancy between two probability distributions, $P$ and $Q$, via the mean embedding distance in a Reproducing Kernel Hilbert Space (RKHS). If two datasets originate from the same distribution, their MMD value should approach zero; conversely, a significantly positive MMD value indicates a distribution mismatch between the two.\newline\indent\setlength{\parindent}{2em}
On the dataset constructed in this study, we formulated the null hypothesis ${H_0}$ : image data from any two different camera nodes follow the same probability distribution. By sampling data from each camera pair and computing MMD statistics directly in the raw pixel space, combined with a permutation test to estimate p-values, the experimental results showed that the data between any two physical scenarios yielded extremely small $p$-values (much lower than the significance level of 0.05). This statistical evidence rejected the null hypothesis ${H_0}$  with high confidence, quantifying that different physical acquisition environments indeed constitute statistically significant distinct domains. This not only reveals the fundamental reason for the generalization failure of traditional single-scenario models in cross-camera tasks but also provides a solid theoretical and empirical basis for the disentangled representation-based adaptation method adopted in this study.
\subsection{Data Preprocessing and Augmentation}
All cropped cow torso images are resized and randomly cropped to 224×224 during training, with random horizontal flipping applied for data augmentation. ImageNet mean and variance are used for normalization.
\subsection{Experimental Platform and Parameter Configuration}
The experiments in this study were conducted on a server equipped with an NVIDIA GeForce GTX 3090 GPU and an Intel i5-13600KF CPU, with software development based on the PyTorch deep learning framework. To validate the model's generalization capability across different camera nodes, a "Leave-One-Camera-Out" strategy was adopted. Specifically, each camera node was alternately selected as the target domain, while the remaining node served as source camera. The average identification accuracy across five tasks was used as the evaluation metric. Comparative baselines covered five ResNet-50-based domain adaptation schemes, all of which completed hyperparameter tuning on the source camera validation set. The proposed method was trained using the SGD optimizer, with an initial learning rate of 0.001, momentum of 0.9, and weight decay of $5 \times {10^{ - 4}}$. The batch size was set to 32, and the training lasted for 40 epochs. In the feature disentanglement module, the dimensions of the latent subspaces ${z_1}$, ${z_2}$, and ${z_4}$ were set to 2, 64, and 10, respectively. The dimension of ${z_3}$, which carries the intrinsic identity information, was set to 256 for most cameras and adjusted to 192 for the field-of-view-limited C4 to enhance feature compactness.

\section{Results and Analysis}
\subsection{Performance Analysis}

\begin{table}[pos=h]
\caption{Quantitative comparison of identification accuracy (\%) on five cross-camera tasks, where $\rightarrow$Ci denotes the evaluation on the $i$-th camera node acting as the target camera.}
\label{tb13}
\begin{tabular*}{\linewidth}{@{} LLLLLLL @{}}
\toprule
Models & $\rightarrow$C1 & $\rightarrow$C2 & $\rightarrow$C3 & $\rightarrow$C4 & $\rightarrow$C5 & Average \\
\midrule
Source-only\citep{he2016deep} & 62.9 & 70.3 & 49.2 & 9.9 & 67.0 & 51.9 \\
DAN\citep{long2015learning} & 66.9 & 84.0 & 58.3 & 47.7 & 86.3 & 68.6 \\
DANN\citep{ganin2015unsupervised} & 51.1 & 81.6 & 58.7 & 40.9 & 83.2 & 63.1 \\
MFSAN\citep{zhu2019aligning} & 53.1 & 82.1 & 61.8 & 68.8 & 84.0 & 70.0 \\
SSG\citep{yuan2022self} & 55.9 & 70.8 & 44.0 & 44.0 & 77.0 & 58.3 \\
iMSDA\citep{kong2023partial} & \textbf{75.8} & 95.1 & 61.2 & 71.3 & 95.8 & 79.8 \\
Ours & 75.5 & \textbf{96.7} & \textbf{88.9} & \textbf{72.6} & \textbf{96.1} & \textbf{86.0} \\
\bottomrule
\end{tabular*}
\end{table}
Table 4 presents the quantitative comparison results for the cross-camera tasks. The proposed feature disentanglement framework achieved an average accuracy of 86.0\%, significantly outperforming the Source-only baseline (51.9\%) and domain adaptation method iMSDA (79.8\%). This advantage stems from the fact that the model regresses to the underlying physical data generation mechanism, actively stripping away imaging interference to extract intrinsic biological features. Compared to statistical distribution alignment methods such as DAN , our algorithm avoids the risk of incorrectly mapping geometric deformations as identity features, thereby enhancing decision robustness under drastic viewpoint changes.\newline\indent\setlength{\parindent}{2em}
Drastic shifts in cross-camera environments represent the primary bottleneck for model generalization. The Source-only model achieved a mere 9.9\% accuracy under the top-down perspective (C4), confirming that patterns learned exclusively from labeled source cameras distributions fail significantly when transferred to complex, unseen target cameras. While methods such as MFSAN  yield marginal improvements through global distribution alignment, their lack of explicit separation between imaging styles and biometric features makes these coarse-grained strategies prone to negative transfer. Specifically, the process of background alignment inadvertently compromises the highly discriminative, fine-grained biometric traits of the cows.
The experimental results further substantiate the superiority of the proposed feature disentanglement paradigm, particularly in the information-constrained $\rightarrow$C3 and $\rightarrow$C4 tasks, where accuracies of 88.9\% and 72.6\% were achieved, respectively. By incorporating the camera index as an auxiliary condition, the model is capable of dynamically filtering out camera-specific interference factors, thereby enabling more precise individual identification in uncontrolled farming environments. This not only validates the effectiveness of the subspace disentanglement approach but also demonstrates the significant application potential of the proposed framework for multi-node collaborative monitoring in smart livestock.

\subsection{Qualitative Insights and Error Case Analysis}
To gain a deeper understanding of the model's decision-making behavior, we conducted a detailed qualitative analysis in addition to overall quantitative accuracy metrics. Fig. 6 aggregates the normalized confusion matrices for the five tasks, aiming to intuitively illustrate the advantages and challenges encountered by the algorithm in practical application scenarios.

\begin{figure}
	\centering
		\includegraphics[scale=.3]{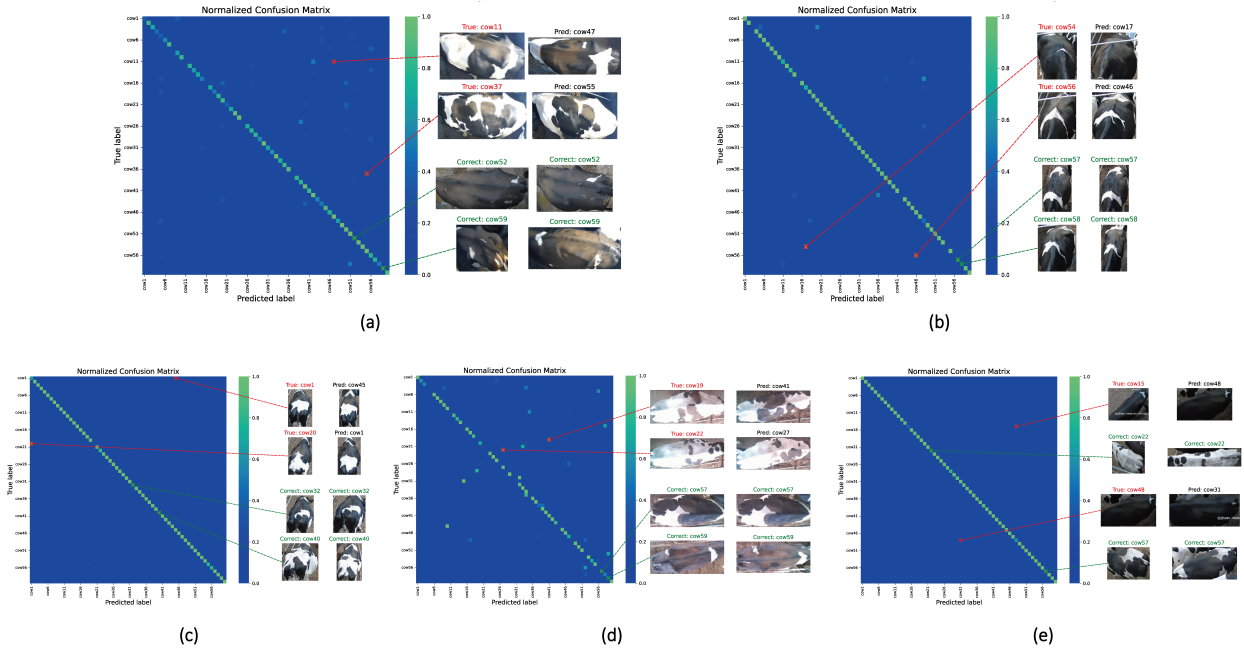}
	\caption{Qualitative performance analysis of the model in cross-camera cow identification tasks. The figure displays normalized confusion matrices and visualizes typical correct (green) and incorrect (red) identification cases.}
	\label{FIG:6}
\end{figure}

The prominent diagonal distribution within the confusion matrices indicates that the model can make precise judgments across the vast majority of scenarios. Even under the extreme top-down perspective of the $\rightarrow$C4 task or the motion blur present in the $\rightarrow$C3 task, the disentangled representation framework successfully penetrates visual noise to capture stable intrinsic biometric features. \newline\indent\setlength{\parindent}{2em}
Simultaneously, failure cases provide valuable insights for future optimization. Analysis indicates that identification errors primarily stem from visual ambiguity caused by extreme physical environments. For instance, the synergistic effect of extreme viewpoints and complex lighting may lead to the deep occlusion of critical biometric patterns or produce severe non-rigid deformations, thereby interfering with the model's predictions. Furthermore, when the observation field of view is restricted and different individuals possess highly similar local patterns, the lack of global information increases the risk of misidentification. In summary, the qualitative case analysis fully validates the effectiveness of the proposed feature disentanglement paradigm in addressing cross-camera scenario shift, providing a robust solution for precision livestock monitoring in uncontrolled environments.

\section{Conclusion and future}
To address the identification failures caused by cross-camera monitoring in precision livestock farming, this study proposes a cross-camera feature disentanglement identification method based on the Subspace Identification Guarantee (SIG) theory. This approach breaks the limitations of traditional statistical alignment methods. It employs a feature disentanglement mechanism, which can explicitly disentangles stable individual identity features that are not related to camera changes from confounding factors such as lighting, perspective, and different imaging styles. This fundamentally enhances the robustness of the identification system in uncontrolled environments. On a real-world livestock dataset covering five distinct observation nodes and various lighting conditions, the proposed framework achieved an average identification accuracy of 86.0\%, significantly outperforming traditional training strategies and mainstream domain adaptation baseline models. The experimental results strongly substantiate that a disentanglement strategy based on underlying physical generative mechanisms can effectively preserve fine-grained discriminative information crucial for individual identification, demonstrating superior adaptability, particularly in the face of drastic geometric deformations induced by camera switching.\newline\indent\setlength{\parindent}{2em}
Although significant progress has been made in cross-camera identification tasks, there remains room for further optimization. Future work will focus on expanding the dataset to include more diverse cow breeds, farming environments, and weather conditions to further validate and enhance the generalization performance of the model. Simultaneously, addressing the practical demands for edge computing in smart livestock through lightweight model modifications to achieve low-cost, real-time terminal deployment represents a research direction of significant application value. Furthermore, the validated feature disentanglement paradigm possesses broad generalization potential and could be extended to other automated monitoring tasks, such as cross-camera cow behavior analysis and body condition scoring. In conclusion, this study not only provides an efficient and robust technical solution for cross-camera cow identification but also lays a theoretical and technical foundation for physically grounded vision analysis in complex farming environments



\bibliography{cas-refs}


\end{document}